\documentclass[10pt, a4paper]{article}
\usepackage{lrec}
\usepackage{graphicx}
\usepackage{tabularx}
\usepackage{soul}

\usepackage{url}

\usepackage{booktabs}
\usepackage{multirow}

\usepackage{amsmath}
\DeclareMathOperator*{\argmax}{arg\,max}
\usepackage[linesnumbered,ruled]{algorithm2e}

\SetKwInput{KwInput}{Input}
\SetKwInput{KwOutput}{Output}

\newcommand{\ACRO}[1]{\textsc{#1}}
\newcommand{\ARRAU}{\ACRO{arrau}}
\newcommand{\CONLL}{\ACRO{conll}}
\newcommand{\CRAC}{\ACRO{crac}}
\newcommand{\PD}{\ACRO{pd}}
\newcommand{\pdfull}{Phrase Detectives}

\newcommand{\DN}{\ACRO{dn}}

\newcommand{\NO}{\ACRO{no}}
\newcommand{\NR}{\ACRO{nr}}
\newcommand{\NLP}{\ACRO{nlp}}
\newcommand{\NP}{\ACRO{np}}
\newcommand{\ONTONOTES}{\ACRO{ontonotes}}
\newcommand{\PREFILTERING}{\ACRO{prefiltering}}

\newcommand{\HYBRID}{\ACRO{hybrid}}
\newcommand{\FINENR}{\ACRO{fine nr}}
\newcommand{\NONR}{\ACRO{no nr}}

\title{A Cluster Ranking Model for Full Anaphora Resolution}
\name{Juntao Yu, Alexandra Uma, Massimo Poesio}
\address{Queen Mary University of London\\
         \{juntao.yu, a.n.uma, m.poesio\}@qmul.ac.uk\\
         }

\abstract{
Anaphora resolution (coreference) systems designed for the {\CONLL} 2012 dataset typically 
cannot handle key aspects of the full anaphora resolution task such as the identification of singletons and of certain types of non-referring expressions (e.g., expletives),  
as these aspects are not annotated
in that corpus.   
However, the recently released {\CRAC} 2018 Shared Task and {\pdfull} ({\PD}) datasets can now be used for that purpose.  
In this paper, we introduce an architecture to simultaneously identify non-referring expressions (including expletives, predicative {\NP}s, and other types)  and build coreference chains, including singletons.   
Our cluster-ranking system uses an attention mechanism to determine the relative importance of the mentions in the same cluster.  Additional classifiers are used to identify singletons and non-referring markables. 
Our contributions  are as follows.
First of all, we report the first result on the {\CRAC} data using system mentions; our result is 5.8\% better than the shared task baseline system, which used gold mentions. Our system also outperforms the best-reported system on {\PD} by up to 5.3\%.
Second, we demonstrate that the availability of singleton clusters and non-referring expressions can 
lead to 
substantially  
improved  performance on non-singleton clusters as well. 
Third, we show  that despite our model not being designed specifically for the {\CONLL} data, 
it  
achieves a very competitive result.% a score equivalent to that of the SoTA system by  \newcite{kantor2019bertee} on that dataset. 
\newline  }

\begin{document}

\maketitleabstract

\section{Introduction}

Anaphora resolution 
is the task of identifying and resolving nominal anaphoric reference to discourse entities \cite{poesio-stuckardt-versley:book}.\footnote{%
%For historical reasons, {\NLP} is mostly focused on the simplified version of this task only concerned with identity reference and known  as coreference resolution since the 1990s \ACRO{muc} shared task.
Some {\NLP} researchers use the term anaphora resolution to refer to pronominal anaphoric reference only, but we use the term in the traditional linguistic and psycholinguistic sense (see \cite{poesio-stuckardt-versley:book} for full discussion).
}
%\footnote{A simplified version of this task is also known in {\NLP} as coreference resolution.}
It is an important aspect of natural language processing and has a substantial impact on  downstream applications such as 
summarization \cite{steinberger-et-al:IPM07,steinberger-et-al:ana_book_summarization}.
Since the {\CONLL} 2012 shared task \cite{pradhan2012conllst},  the {\ONTONOTES} corpus has been the dominant resource in research on identity anaphora resolution (coreference) \cite{fernandes-et-al:CL14,bjorkelund2014learning,martschat&strube:TACL15,clark2015entity,clark2016deep,clark2016improving,lee2017end,lee2018higher,kantor2019bertee,joshi2019bert,joshi2019spanbert}.
But {\ONTONOTES} has a number of limitations.
An often mentioned limitation
is that singletons are not annotated \cite{DeMarneffe:2015:MLD:2831407.2831417,chen-EtAl:2018:EMNLP2}. 
A less discussed, but still crucial, limitation is that 
although some types of non-referring expressions are marked in {\ONTONOTES}, in particular predicative ones (\textit{a policeman} in \textit{John is a policeman}), 
other types are not, 
such as expletives, 
meaning that in \textit{It rained}, \textit{It} 
is not 
considered a markable. 
As a consequence, 
systems optimized for 
{\ONTONOTES} are only evaluated on non-singleton coreference chains; their performance at identifying  singletons,  and 
distinguishing them from expletives, is not evaluated.

But the decision to interpret \textit{it} as referring or non-referring \cite{Uryupina-et-al:ana_book_non_referring,versley-EtAl:2008:PAPERS,bergsma-lin-goebel:2008:ACLMain,Bergsma:2011:nada,hardmeier-EtAl:2015:DiscoMT}  is a key aspect of pronoun interpretation--for instance, for the purposes of machine translation \cite{GUILLOU16.327}--so systems trained on {\ONTONOTES} have had to adopt a variety of workarounds.
These  limitation of {\ONTONOTES} have however been 
corrected
in a number of corpora, including \ACRO{ancora} for Spanish \cite{taule-et-al:2008:ancora}, \ACRO{tuba-d/z} for German \cite{telljohann2006stylebook}, and, for English, {\ARRAU}  \cite{uryupina-et-al:NLEJ}, 
which 
was used as dataset for the {\CRAC} 2018 shared task 
\cite{poesio2018crac}, 
and {\pdfull} ({\PD}) \cite{poesio-etal-2019-crowdsourced}. 

The first contribution of this paper is the development of a system able to perform both coreference resolution and identification of non-referring markables and singletons, using the {\CRAC} 2018 shared task and {\PD} datasets. 
On {\CRAC}, our model achieves a {\CONLL} score of 77.9\% on coreference chains, and an  F1 score of 76.3\% on non-referring expressions identification. 
This is, to the best of our knowledge, the first modern result on the {\CRAC} data using system mentions. 
Our {\CONLL} score is even 5.8\% higher than the baseline result on this dataset, 72.1\% obtained by \cite{poesio2018crac} using gold mentions. 
On {\PD}, our model outperforms the best-performing system by up to 5.3\%.

Our second contribution is a novel and competitive cluster ranking architecture for anaphora resolution\footnote{The code is available at 
https://github.com/juntaoy/dali-full-anaphora}.
%http://AnonymousLink}. 
Current coreference models can be classified either as mention pair models \cite{soon-et-al:CL01}, in which connections are established between mentions, or entity mention models, in which mentions are directly linked to  entities / coreference chains  \cite{luo-et-al:ACL04,rahman&ng:JAIR11}.  
The mention pair models are simpler in concept and easier to implement, so many SoTA systems are exclusively based on mention ranking  \cite{wiseman2015learning,clark2016deep,lee2017end}. 
But it  has long been known that  entity-level information is important for coreference \cite{luo-et-al:ACL04,poesio-stuckardt-versley:book} so many systems attempted to explore features beyond those of mention pairs \cite{bjorkelund2014learning,clark2015entity,clark2016improving,lee2018higher,kantor2019bertee,joshi2019bert,joshi2019spanbert}.  
However, those systems are usually much more complex than their mention ranking counterpart, since entity features are introduced in addition to their mention ranking part. 
Consider the \newcite{lee2018higher} system, for instance: the full system has 9.6 million trainable parameters in total, 
which is double the number of 
the mention ranking part of the system (4.8M parameters). In this work, we demonstrate that it is possible to achieve SoTA results by cluster ranking alone, i.e. by linking mentions directly to the entities. As a result, our model is less complex than the existing entity-level models \cite{lee2018higher,kantor2019bertee} using similar mention representations. 
Our model uses only 4.8M trainable parameters without increasing the complexity of a mention ranking model. 
Furthermore, our model is fast to train; we show that a cluster ranking model can be significantly sped up by training on oracle clusters\footnote{The oracle clusters are created from system mention using gold cluster information.}.

The key intuitions behind the proposed approach are (i) that cluster representations are crucial to the success of a cluster ranking system, and (ii) that a key property of these representations is that they should capture the fact that mentions in a cluster are not equally important. In particular, it is well-known that the mentions introducing an entity are generally more informative (e.g., \textit{the president of ACME, John Smith}) whereas subsequent mentions tend to employ reduced forms (e.g., \textit{Mr. Smith, he}) \cite{ariel:90}. This motivates the use of cluster representations capable of preserving the greater importance of earlier mentions. 
Our approach captures this mention importance by using attention scores for the mentions in a cluster and combining the mention representations according to their attention scores. We then investigate the effect of the cluster histories by including all the history of the clusters as candidate assignments to the mentions. 
The resulting system, besides achieving the new SoTA on the {\CRAC} dataset (whether including and excluding non-referring expressions and singletons),
achieves {\CONLL} scores equivalent to the current SoTA system not fine-tuned on BERT \cite{kantor2019bertee} on {\CONLL} data as well (in which non-referring expressions and singletons are not annotated).

Our third and final contribution is the finding that training 
our system
on annotations of singleton mentions and non-referring expressions 
enhance its performance on non-singleton coreference chains. 
By evaluating our system on the {\CRAC} data we show that gains of up to 1.4 percentage points on non-singleton coreference chains can be achieved by training the model with additional singleton mentions and non-referring expressions.

\section{System architecture}
Anaphora 
resolution is the task of identifying the referring mentions in a text and assigning those mentions to disjoint clusters such that mentions in the same cluster refer to the same entity. 
The first subtask of anaphora resolution is  mention detection, i.e., extracting candidate mentions from the document. 
Until recently, 
most  coreference systems 
selected mentions 
prior to coreference resolution 
via heuristic methods
often
based on parse trees \cite{bjorkelund2014learning,clark2015entity,clark2016deep,clark2016improving,wiseman2015learning,wiseman2016learning}. \newcite{lee2017end}  introduced a neural network %based 
approach for joint mention detection and coreference resolution, obtaining the best performing system at the time. 
The system was further extended by \newcite{lee2018higher}, \newcite{kantor2019bertee}, 
\newcite{joshi2019bert} and \newcite{joshi2019spanbert}, the current SoTA on the {\CONLL} data set.

Our model is also a joint system that predicts  mentions and assigns them to the clusters jointly. For a given document $D$ with $T$ tokens, we define all possible spans in $D$ as $N_{i=1}^{I}$ where $I = \frac{T(T+1)}{2}$, $s_i,e_i$ are the start and the end indices of $N_i$ where $1 \leq i \leq I$. 
The task for a joint system is to partition %assign 
all the spans ($N$) into a sequence of clusters $(C^{m})\ _{m=1}^{M}$ such that every mention in a specific cluster $C^{m}$ refers to the same entity. 
Let $C_i$ be the partially completed clusters up to span $N_i$. 
The set of possible assignments for $N_i$ is defined as all the clusters up to the previous span ($C_{i-1}$) and a special label $\epsilon$. 
The $\epsilon$ is used for three situations: a span is not a mention, or is a non-referring expression, or is the first mention of a cluster. 

\begin{algorithm}[t]
\small
\SetAlgoLined
\KwInput{$(\hat{N}_i^*, s_m(i),s_\epsilon(i),\beta(i))_{i=1}^{\lambda T}$}
\KwOutput{$C_{\lambda T}$}
$m=0; C_0=\{\}; s_{c_0}=\{\}$\;
\For{$i : 1..\lambda T $ }{
$ \textsc{Tmp} \leftarrow  s_\epsilon (i)$\;
\For{$j :1..m$}{
%calculate $s_{mc}(i,j)$\; 
$\textsc{Tmp} \leftarrow s_m(i) + s_c(j) + s_{mc}(i,j)$
}
$b = \argmax \textsc{Tmp}$\;
\uIf{$b==\epsilon$}{
$C_i = C_{i-1} \cup \{\hat{N}^*_i\}$\;
$s_{c_i} = s_{c_{i-1}} \cup s_m(i)$\;
$m = m+1$\;
}\Else{
%calculate $a_{i}^{b}(m)\ (m\ \in C_{i-1}^{b} \cup \hat{N}_i ) $\;
$C_i^b = \sum_{m \in C^b_{i-1} \cup \hat{N}_i} a_{i-1}^{b}(m) \cdot \hat{N}_m^*$\;
$s_{c_i}(b) = \sum_{m \in C^b_{i-1} \cup \hat{N}_i} a_{i-1}^{b}(m) \cdot s_m(m)$\;
}
}
\caption{Cluster ranking algorithm.}\label{algorithm:att}
\end{algorithm}
%\vspace{-15pt}

\subsection{Mention Representation}
%In this work, 
We use 
a 
mention representation 
based on those in 
\cite{lee2018higher,kantor2019bertee}.
Our system represents a candidate span with the outputs of a BiLSTM, encoding
the sentences in a document %are encoded 
from both directions to obtain a representation for each token in the sentence. 
The BiLSTM takes as input the concatenated embeddings ($(x_t)_{t=1}^{T}$) of both word and character levels.
For word embeddings,  GloVe \cite{pennington2014glove} and BERT \cite{devlin2019bert} embeddings are used. 
Character embeddings are learned by a convolution neural networks (CNN) during training. 
The tokens are represented by concatenated outputs from the forward and the backward LSTMs. 
The token representations $(x^{*}_t)_{t=1}^{T}$  are used together with head representations ($h^*_i$) to represent candidate spans ($N^*_i$). The $h^*_i$ of a span is obtained by applying an attention over its token representations ($\{x^{*}_{s_i}, ..., x^{*}_{e_i}\}$), where $s_i$ and $e_i$ are the indices of the start and the end of the span respectively. Formally, we compute $h^*_i$, $N^*_i$ as follows:

\vspace{-10pt}
\small
\begin{align*}
\alpha_t &= \textsc{ffnn}_{\alpha}([x^{*}_{t}, \phi(t)])\\
a_{i,t} &= \frac{exp(\alpha_t)}{\sum^{e_i}_{k=s_i} exp(\alpha_k)}\\
h^*_i &= \sum^{e_i}_{t=s_i} a_{i,t} \cdot x_t\\
N^*_i &= [x^*_{s_i}, x^*_{e_i},h^*_i,\phi(i)]
\end{align*}
\normalsize

where $\phi(t), \phi(i)$ are the cluster position and span width feature embeddings respectively.

To make the task computationally tractable, our model only considers the spans up to a maximum length of $l$, i.e. $e_i -  s_i < l, \ (s_i,e_i) \in N$. Further pruning is applied before feeding the candidate mentions to the coreference resolver. 
The top ranked $\lambda T$ spans are selected from $lT$ candidate spans ($\lambda < l$) by a scoring function $s_m$. where:

\small
$$s_m(i) =\textsc{ffnn}_{m} (N^*_i) $$
\normalsize

The top $\lambda T$ selected spans are required \textit{not} to be partially overlap, i.e. there is no such cases that $s_i < s_j \leq e_i < e_j$ or $s_j < s_i \leq e_j < e_i$. The nested spans are not affected by this constrains since they are not partially overlap.

\subsection{The Cluster Ranking Model}
Let $(\hat{N}_i)_{i=1}^{\lambda T}$ denote the top ranked $\lambda T$ candidate mentions selected by the mention detector after pruning. 
The model builds the clusters $(C^m)_{m=1}^M$ by visiting $\hat{N}_i$ in  text order and assigning them a cluster in the case  $i \neq \epsilon$, or creating a new cluster if $i = \epsilon$. 
Let $C_{i}$ be the partial clusters consisting of up to $i_{th}$ mentions, %($\{\hat{N}_1, ..., \hat{N}_i\}$), 
and $c_i$  the cluster assigned to $\hat{N}_i$. 
The task of our cluster ranking model is to output $\hat{C}$ that maximises the score of the final clusters:

\small
$$
\hat{C}\ =\ \argmax_{c_1,...,c_{\lambda T}}\sum_{i=1}^{\lambda T} s(i, c_i)
$$
\normalsize

where $s(i,j)$\footnote{We follow \newcite{lee2018higher} and use $i$ to indicate the anaphor and $j$ for the antecedent.} is a scoring function between a mention $N_i$ and a set of possible assignments $j \in \{\epsilon, C^m_{i-1}\}$:

\small
$$
s(i,j) = \Bigg\{
  \begin{tabular}{ll}
  $s_{\epsilon}(i)$& $j=\epsilon$ \\
  $s_m(i) + s_c(j)+ s_{mc}(i,j)$& $j\neq \epsilon$
  \end{tabular}
$$
\normalsize

and $s_\epsilon(i)$ is the probability that $\hat{N}_i$ does \textit{not} belongs to any of the previous clusters $C^m_{i-1}$. To use a scoring function for $\epsilon$ instead of a constant 0 (used by \newcite{lee2018higher}) gives us the flexibility to extend the function for handing more detailed types of $\epsilon$, such as non-referring.     $s_m(i)$ is the mention score that has been used to rank the candidate mentions. $s_c(j)$ is the cluster score computed from the mention scores that belongs to the cluster. $s_{mc}(i,j)$ is a pairwise score between $i_{th}$ mention $\hat{N}_i$ and $j_{th}$ partial cluster of $C^j_{i-1}$. 
To implement the cluster ranking model we use an attention function $a(m)$ %mechanism 
\cite{bahdanau2014neural} to assign an importance to each of the mentions. 
We compute the cluster score $s_c(j)$ and the cluster representation ($C^{j^*}_{i-1}$) (for computing $s_{mc}(i,j)$), by mention scores/representations and with consideration of  mention 
importance.  % salience
More precisely, we compute the scores as follows:

\small
\vspace{-10pt}
\begin{align*}
s_\epsilon(i) &= \textsc{ffnn}_\epsilon (\hat{N}^*_i) \\
s_m(i) &= \textsc{ffnn}_{m} (\hat{N}^*_i) \\
\beta(i) &= \textsc{ffnn}_{\beta}([\hat{N}^{*}_{i},\phi(i_\beta)])\\
%\end{align*}
%\begin{align*}
a_{i-1}^{j} (m) &= \frac{exp(\beta(m))}{\sum_{k \in C^j_{i-1}} exp(\beta(k))}\\
s_{c_{i-1}}(j) &= \sum_{m \in C^j_{i-1}} a_{i-1}^{j}(m) \cdot s_m(m)\\
C^{j^*}_{i-1} &= \sum_{m \in C^j_{i-1}} a_{i-1}^{j}(m) \cdot \hat{N}_m^*\\
F^*_{(i,j)} &= [\hat{N}^*_i, C^{j^*}_{i-1}, \hat{N}^*_i \circ C^{j^*}_{i-1},\phi(i,\hat{j}),\phi(j)]\\
s_{mc}(i,j)&=\textsc{ffnn}_{mc}(F^*_{(i,j)})
\end{align*}
\normalsize

Both $s_c(j)$ and $C^{j^*}_{i-1}$ are updated each time a cluster is expanded. 
$\phi(i_\beta)$ is the position embeddings that indicates the position of a mention in the cluster. $\phi(i,\hat{j})$ is a small set of features between the $\hat{N}_i$ and the newest mention $\hat{N}_{\hat{j}}$ of the cluster. 
We used the same features as \newcite{lee2018higher}:
these include genre, speaker (boolean, same or not) and distance (between $i$ and $\hat{j}$) features. $\phi(j)$ is %the 
cluster size, 
a common entity-level feature \cite{bjorkelund2014learning}. 
The size is assigned into buckets according to its value. We use the buckets of \newcite{bjorkelund2014learning}, assigning the values in 8 buckets ([1,2,3,4,5-7,8-11,12-19,20+]).
The pseudo-code of our model is shown in Algorithm \ref{algorithm:att}.%\footnote{We also evaluated an alternative approach, in which the clusters are encoded by a LSTM. However the LSTM approach resulted in a lower accuracy than the attention approach in this evaluation.} 
\footnote{We do \textit{not} use  coarse-to-fine pruning or higher-order inference, unlike \newcite{lee2018higher} and \newcite{kantor2019bertee}.  
We found coarse-to-fine pruning  does \textit{not} improve our model when compared with simpler distance pruning. 
As for higher-order inference, our system  already has access to the entity-level information by default, hence it is not necessary.}

\subsection{Cluster History}

One of the advantages of the mention ranking model is that the correct cluster can be built by attaching the active mention to any of the antecedents in the correct cluster. 
This reduces the complexity of the task as there are multiple correct links. 
By contrast, in a standard cluster ranking model, only one correct cluster can be chosen. 
In order to make 
multiple links possible in our
cluster ranking system, we extended our
model by including all cluster histories (\ACRO{ch});
this maximises the chance of choosing 
the correct clusters.  
(We make sure a mention is always attached to the latest version of the cluster by including
an additional pointer linking  
every cluster history 
to the latest version of the cluster.)
This makes the model slightly more similar to a mention ranking model; 
however, there is still a fundamental difference, as we use  cluster representations instead of  mention representations.
We replace the line 13 and 14 of Algorithm \ref{algorithm:att} 
to get the model 
that includes  cluster histories:

\vspace{-10pt}
\small
\begin{align*}
b &= \textsc{Latest}(b)\\
C_i &= C_{i-1} \cup  \sum_{m \in C^b_{i-1} \cup \hat{N}_i} a_{i-1}^{b}(m) \cdot \hat{N}_m^*
\end{align*}
\begin{align*}
s_{c_i} &= s_{c_{i-1}}\cup \sum_{m \in C^b_{i-1} \cup \hat{N}_i} a_{i-1}^{b}(m) \cdot s_m(m)\\
m &= m+1
\end{align*}
\normalsize

where $\textsc{Latest}(b)$ is a function to find the latest version of the cluster $b$.

\subsection{Identifying Non-Referring Expressions}

To add  non-referring expressions identification, we extend %the
$\epsilon$ into multiple classes:
{\NO} for non-mention, 
{\NR} for non-referring and 
{\DN} for discourse new, including singletons

\small
$$
s_{\epsilon}(i) = \Bigg\{
  \begin{tabular}{ll}
  $s_{no}(i)$& \NO \\
  $s_{nr}(i) + s_m(i)$&  \NR\\
  $s_{dn}(i) + s_m(i)$ & \DN
  \end{tabular}
$$
\normalsize

Several non-referring types are annotated in the {\ARRAU} corpus: in addition to expletives, there are also predicative {\NP}s (e.g., \textit{a policeman} in \textit{John is \underline{a policeman}}), non-referring quantifiers (e.g.,\textit{nobody} in \textit{I see \underline{nobody} here} ) \cite{karttunen:76}, idioms (e.g., \textit{her hand} in \textit{He asked her for \underline{her hand}}), etc.
As we will see, the basic {\NR} classifier can be  extended 
to do 
a fine-grained classification of non-referring expressions. 

By distinguishing 
`non-mentionhood' %non-mentions 
from 
non-anaphoricity %discourse new 
the system naturally resolves  singletons (i.e. the clusters with a size of one).   
Non-referring expressions are usually  filtered before building the coreference chains, e.g. in  \ACRO{mars}  \cite{mitkov2002new}; 
we will call this {\PREFILTERING} approach. 
In the {\PREFILTERING} approach, the system removes the markables identified as non-referring expressions from further processing once they have been identified. To be more specific, we replace line 8 of algorithm \ref{algorithm:att} with:

\vspace{-10pt}
\small
\begin{align*}
\textbf{if}&\ b=={\NO}\ \textbf{or}\ b=={\NR}\ \textbf{then}\\
    &\ \ \ C_i = C_{i-1};\ \ s_{c_i}= s_{c_{i-1}};\ \ m = m;\\
\textbf{else if}&\ b=={\DN}\ \textbf{then}
\end{align*}
\normalsize

The {\PREFILTERING} approach is aggressive,  which might have a negative effect on results if referring expressions have been filtered incorrectly. 
We also tried therefore a second approach: only do prefiltering when the non-referring expressions classifier has high confidence (when the classifier has a softmax score above a heuristic threshold $t\ (0 \leq t \leq 1)$). The softmax score is calculated between previous clusters and classes in $\epsilon$ (i.e. \textsc{Tmp} in algorithm \ref{algorithm:att}). 
If the score is below this threshold,
non-referring expressions are identified after (postfiltering) forming the clusters (we call this {\HYBRID} approach). During postfiltering, candidates that are classified as non-referring markables with lower confidence and are not part of clusters are included as additional non-referring markables.

\begin{table}[t]
    \centering
    \small
    \begin{tabular}{l l}
    \toprule
    \bf Parameter & \bf Value \\
    \midrule
    BiLSTM layers/size/dropout &3/200/0.4\\
    %BiLSTM size &200\\
    %BiLSTM dropout &0.4\\
    FFNN layers/size/dropout & 2/150/0.2\\
    %FFNN size & 150\\
    %FFNN dropout & 0.2\\
    CNN filter widths/size& [3,4,5]/50\\
    %CNN filter size&50\\
    Char/GloVe/Feature embedding size&8/300/20\\
    %GloVe embedding size&300\\
    BERT embedding size/layer&1024/Last 4\\
    %BERT embedding layer&Last 4\\
    %Feature embedding size&20\\
    Embedding dropout & 0.5\\
    Max span width ($l$)&30\\
    Max num of clusters&250\\
    Mention/token ratio ($\lambda$) &0.4\\
    Optimiser & Adam (1e-3)\\
    %Learning rate & 1e-3\\
    %Decay rate&0.999\\
    %Decay frequency&100\\
    Training step & 200K\\
    \bottomrule
    \end{tabular}
    \caption{Hyperparameters for our models.}
    \label{table:config}
\end{table}

\begin{table*}[t]
\centering
\small
\begin{tabular}{l l l l l l l l l l l l}
\toprule
 & \multirow{2}{*}{Models} & \multicolumn{3}{l}{MUC} & \multicolumn{3}{l}{B$^3$} & \multicolumn{3}{l}{CEAF$_{\phi_4}$} & \multirow{2}{*}{\begin{tabular}[c]{@{}l@{}}Avg.\\ F1\end{tabular}} \\ \cmidrule{3-11}
 & & P & R & F1& P & R & F1& P& R  & F1 & \\ \midrule
\multirow{4}{*}{\begin{tabular}[c]{@{}l@{}}Singletons \\included\end{tabular}}
&{\PREFILTERING}&75.5&\bf 79.0&77.2&75.9&\bf 80.7&78.2&75.2&77.3&76.2&77.2\\ %\cmidrule{2-12}
&{\HYBRID}&\bf 77.9&78.5&\bf 78.2&\bf 77.4&80.3&\bf 78.8&\bf 75.4&78.1&\bf 76.8&\bf 77.9
\\ %\cmidrule{2-12}
&{\FINENR}&76.7&77.3&77.0&76.8&79.7&78.2&74.9&78.0&76.4&77.2\\ \cmidrule{2-12}
& \newcite{lee-et-al:CL13}* &72.1&58.9&64.8&77.5&77.1&77.3&64.2&\bf 88.1&74.3&72.1\\
\midrule 
\multirow{5}{*}{\begin{tabular}[c]{@{}l@{}}Singletons \\excluded\end{tabular}}&
{\PREFILTERING}&75.5&\bf 79.0&77.2&67.0&\bf 73.0&69.9&67.1&\bf 65.1&66.1&71.1\\%\cmidrule{2-12}
&{\HYBRID}&\bf 77.9&78.5&\bf 78.2&\bf 69.2&71.8&\bf 70.4&\bf 69.5&63.8&\bf 66.5&\bf 71.7\\%\cmidrule{2-12}
&{\FINENR}&76.7&77.3&77.0&68.0&70.7&69.3&66.6&64.2&65.4&70.6\\
&{\NONR}&76.7&77.0&76.8&68.7&69.7&69.2&66.1&63.8&64.9&70.3\\ \cmidrule{2-12}
&\newcite{lee-et-al:CL13}* &72.3& 58.9& 64.9&67.9& 48.5& 56.5&54.2& 53.0&53.6&58.3\\
\bottomrule
\hline
\end{tabular}
\caption{\label{table:finalcracaccuracy} The comparison between our models and the SoTA system on the {\CRAC} test set. * indicates systems evaluated on the gold mentions.}
\end{table*}

\subsection{Learning}
To train a cluster ranking model on system clusters is challenging, as we need to find a way to learn from the partially correct clusters. It is also slow, as the system processes one mention at a time, hence cannot benefit largely from parallel computing.
The solution we adopted was training the model on oracle clusters.
This 
is simpler and faster, since the clusters for one training document can be created before computing more heavy stuff, e.g. the cluster scores $s_c(j)$ and pairwise scores $s_{mc}(i,j)$. 
More precisely, we create the oracle clusters during the training using gold cluster ids; system mentions belonging to the same gold clusters are grouped.  
This is much faster 
than training the model
on the system mentions directly, since training on the system mentions requires
computing scores for each mention separately. 
In a preliminary experiment, we discovered that by training on 
oracle clusters 
we obtain not only a better {\CONLL} score, but also a 
fivefold speedup
compared with the model trained on the system mentions directly.\footnote{We train both approaches on the {\CONLL} data for 200K steps on a GTX 1080Ti GPU. It takes 16 and 80 hours to train a model on oracle and system mentions respectively.}

As a 
loss function, we 
optimize on the marginal log-likelihood of all the clusters that contain mentions from the same gold cluster \textsc{gold}$(i)$ of $\hat{N}_i$.
Formally,

\small
$$log \prod_{i=1}^{\hat{N}}\sum_{\hat{c} \in C_{i-1} \cap \textsc{gold}(i)} P(\hat{c})$$
\normalsize

In case $C_{i-1}$ does not contain any mention from $\textsc{gold}(i)$ or $\hat{N}_i$ does not belongs to a gold cluster, we set \textsc{gold}$(i) = \{\epsilon\}$. For our model to have more than one class in $\epsilon$, the \textsc{gold}$(i)$ is set to the relevant classes ({\NO},{\NR} or {\DN}).

\section{Data and Hyperparameters}

For full anaphora resolution, our 
primary evaluation dataset was 
the {\CRAC} 2018 corpus \cite{poesio2018crac}. 
In addition, we evaluated our model on the {\PD} corpus, also containing expletives.
Finally, we  evaluated our model on the {\CONLL} 2012 English corpora \cite{pradhan2012conllst} to compare its performance with the SoTA on the {\CONLL} task. %(identifying coreference chains excluding singletons, no non-referring expression identification). 

The {\CRAC} Task 1 dataset is based on the \ACRO{rst} portion of the {\ARRAU} corpus \cite{uryupina-et-al:NLEJ}.
% {\ARRAU} includes texts from three very distinct domains: 
% news (the \ACRO{rst} subcorpus), 
% dialogue,% (the \ACRO{trains} subcorpus), 
% fiction.% (the \ACRO{pear} stories),
% %and non-fiction,% (the \ACRO{gnome} corpus).
The %{\ARRAU} 
annotation scheme specifies the annotation of referring expressions (including singletons) and non-referring expressions; split antecedent plurals, generic references, and discourse deixis are annotated, as well as bridging references. The \ACRO{rst} portion of {\ARRAU} consists of news texts (1/3 of the \ACRO{penn} Treebank), 
with 228,000 tokens and 72,000 mentions.

{\PD} is a constantly growing corpus collected using the annotation game {\pdfull} \cite{poesio-etal-2019-crowdsourced}. The corpus was annotated by players and then aggregated by an aggregating method to create a silver standard corpus. 
Both singletons and non-referring markables are annotated. 
We used the latest release of the corpus, consisting  of 542 documents, 408,000 tokens and 108,000 mentions.\footnote{https://github.com/dali-ambiguity/Phrase-Detectives-Corpus-2.1.4}.

The {\CONLL} datasets are the standard datasets 
for coreference. 
The English {\CONLL} corpus consists of 
3493 %2802, 342, and 348 
documents 
%for the train, development and test sets respectively, 
for a total of 1.6M tokens and 194,000 mentions. 

We use the official {\CONLL} 2012 scorer %scoring script 
to score our predictions when evaluating without singletons and non-referring markables, 
and the official {\CRAC} 2018 scorer %scoring script
\cite{poesio2018crac} to evaluate other cases. 
The {\CRAC} 2018 Extended Scorer is an extension of the {\CONLL} 2012 official scorer developed by Nafise Moosavi that can handle singletons and non-referring markables. 
The Extended Scorer is identical to %the same as 
the {\CONLL} scorer when evaluating without singletons and non-referring markables, but also reports P, R and F1 values for non-referring markables when those are considered. 
Following standard practice, we report recall, precision, and F1 scores for MUC, B$^3$ and CEAF$_{\phi_4}$ and the average F1 score of those three metrics. Besides, we report the F1 score for non-referring when needed.

For our experiments, we use the same maximum span width ($l=30$), number spans per tokens ($\lambda=0.4$) and most of the network parameters as \newcite{lee2018higher} and \newcite{kantor2019bertee}. 
The details are %A detailed configurations can be find 
in Table \ref{table:config}.

\section{Results and Discussions}
\label{sec:results}

\subsection{Evaluation on the {\CRAC} data set}

We first compared the two proposed approaches for using non-referring expressions, {\PREFILTERING} and {\HYBRID}.  
For our {\HYBRID} model, we set the threshold ($t$) to 0.5 after tuning on the development set. Table \ref{table:finalcracaccuracy} shows the results of our models on the {\CRAC} test set. 
As expected, the {\HYBRID} model,  using a less greedy pruning, achieved better F1 scores on all three coreference metrics. 
In terms of the non-referring scores (see Table \ref{table:cracnraccuracy}), the {\PREFILTERING} approach has better recall and F1 score, while the {\HYBRID} approach has better precision. 
We hypothesize this is mainly because the {\PREFILTERING} approach generates more non-referring expressions due to its greedy pruning--i.e., the {\PREFILTERING} approach keeps all the candidate non-referring markables once they are identified--while the {\HYBRID} approach  favours  the coreference clusters for non-referring markables fall below the threshold. 
The {\HYBRID} approach has a better overall performance according to our weighed F1 scores (0.85 * \ACRO{coref} F1 + 0.15 * {\NR} F1) The weights are determined by the proportion of the referring and non-referring markables in the corpus.

\begin{table}[t]
\centering
\small
\begin{tabular}{l l l l}
\toprule
Models& P&R&F1\\\midrule
{\PREFILTERING}&76.6&74.5&75.5\\ %\cmidrule{2-5}
{\HYBRID}&78.0&72.4&75.1\\% \cmidrule{2-5}
{\FINENR}&77.0&75.5&76.3\\
\bottomrule
\end{tabular}
\caption{\label{table:cracnraccuracy} The  scores for non-referring expressions of our models on the {\CRAC} test set.}
\end{table}

\begin{table}[t]
\centering
\small
\begin{tabular}{l l l l}
\toprule
{\NR} types& P&R&F1\\ \midrule
Expletive&93.8&100.0&96.8 \\%tp:45, fp:3, fn:0
Predicate&77.6&75.2&76.4 \\%tp:479, fp:138, fn:158
Quantifier&65.0&64.7&64.9 \\%tp:167, fp:90, fn:91
Coordination&77.5&82.0&79.7 \\%tp:283, fp:82, fn:62
Idiom&77.0&55.9&64.8 \\%tp:57, fp:17, fn:45
\bottomrule
\end{tabular}
\caption{\label{table:finenraccuracy} The scores of our models on the fine-grained non-referring types.}
\end{table}

\begin{table*}[t]
\centering
\small
\begin{tabular}{l l l l l l l l l l l l}
\toprule
 & \multirow{2}{*}{Models} & \multicolumn{3}{l}{MUC} & \multicolumn{3}{l}{B$^3$} & \multicolumn{3}{l}{CEAF$_{\phi_4}$} & \multirow{2}{*}{\begin{tabular}[c]{@{}l@{}}Avg.\\ F1\end{tabular}} \\ \cmidrule{3-11}
 & & P & R & F1& P & R & F1& P& R  & F1 & \\ \midrule
\multirow{2}{*}{\begin{tabular}[c]{@{}l@{}}Singletons \\included\end{tabular}}
&\newcite{poesio-etal-2019-crowdsourced}&79.3&72.5&75.7&72.1&69.3&70.7&70.5&73.2&71.8&72.7\\ 
&Our model&\bf 81.9&\bf 76.4&\bf 79.1&\bf 74.9&\bf 73.7&\bf 74.3&\bf 72.2&\bf 75.1&\bf 73.6&\bf 75.7\\ 
\midrule 
\multirow{2}{*}{\begin{tabular}[c]{@{}l@{}}Singletons \\excluded\end{tabular}}&
\newcite{poesio-etal-2019-crowdsourced}&79.3& 72.5& 75.7&58.3&52.4&55.2&58.3&49.5&53.5& 61.5\\
&Our model&\bf 81.9&\bf 76.4&\bf 79.1&\bf 64.7&\bf 61.0&\bf 62.8&\bf 62.9&\bf 54.8&\bf 58.6&\bf 66.8\\
\bottomrule
\hline
\end{tabular}
\caption{\label{table:finalpdaccuracy} The comparison between our models and the SoTA system on the {\PD} test set.}
\end{table*}

\begin{table*}[t]
\centering
\small
\begin{tabular}{r l l l l l l l l l l l}
\toprule
 &\multirow{2}{*}{Models} & \multicolumn{3}{l}{MUC} & \multicolumn{3}{l}{B$^3$} & \multicolumn{3}{l}{CEAF$_{\phi_4}$} & \multirow{2}{*}{\begin{tabular}[c]{@{}l@{}}Avg.\\ F1\end{tabular}} \\ \cmidrule{3-11}
 & & P & R & F1& P & R & F1& P& R  & F1 & \\ \midrule
%\multicolumn{11}{c}{Models use Context Independent Embeddings}\\\midrule
%\newcite{clark2016improving}&\bf 79.9& 69.3& 74.2& \bf 71.0& 56.5& 63.0& 63.8& 54.3& 58.7& 65.3\\
Context&\newcite{clark2016deep}&79.2& 70.4& 74.6& \bf 69.9& 58.0& 63.4& 63.5& 55.5& 59.2& 65.7\\
Independent&\newcite{lee2017end}&78.4& 73.4& 75.8& 68.6& 61.8& 65.0& 62.7& \bf 59.0& 60.8& 67.2\\ 
Embeddings&\newcite{zhang2018acl}&\bf 79.4& \bf 73.8& \bf 76.5& 69.0& \bf 62.3& \bf 65.5& \bf 64.9& 58.3& \bf 61.4& \bf 67.8\\\midrule
%\multicolumn{11}{c}{Models use Pre-trained Context Dependent Embeddings}\\\midrule
Pre-trained&\newcite{lee2018higher}& 81.4 &79.5 & 80.4 & 72.2 &69.5 & 70.8 & 68.2 & 67.1 & 67.6 & 73.0\\
Contextual&\newcite{kantor2019bertee}&82.6 & \bf 84.1& \bf 83.4& 73.3&  \bf 76.2& \bf 74.7& \bf 72.4&  \bf 71.1& \bf 71.8& \bf 76.6\\
Embeddings&Our model&\bf 82.7&83.3&83.0&\bf 73.8&75.6&\bf 74.7&72.2&71.0&71.6&76.4\\ \midrule
%\multicolumn{11}{c}{Models Fine-tuned on BERT}\\\midrule
Fine-tuned&\newcite{joshi2019bert}& 84.7& 82.4&  83.5&  76.5& 74.0&  75.3& 74.1 &69.8 & 71.9 & 76.9\\
on BERT&\newcite{joshi2019spanbert}&\bf 85.8&\bf 84.8&\bf 85.3&\bf 78.3&\bf 77.9&\bf 78.1&\bf 76.4&\bf 74.2&\bf 75.3&\bf 79.6\\
\bottomrule
\end{tabular}
\caption{\label{table:finalaccuracy} Comparison between our models and the top performing systems on the {\CONLL} test set.}
\end{table*}
\textbf{Fine-grained Non-referring}

We further extended the basic {\NR} classifier to recognise the more fine-grained classification of non-referring expressions annotated in the {\CRAC} dataset
by configuring % We configured  
our {\HYBRID} model to learn from the fine-grained types ({\FINENR}). 
Our model does very well 
on resolving expletives (96.8\% F1) 
and achieves 76 - 80\% F1 score on predicates and coordinations, 
but has a lower F1 score of around 65\% on recognising non-referring quantifiers and idioms. 
We also compared this model with the other models to dealing with non-referring expressions by collapsing the classifications it produces (Table \ref{table:cracnraccuracy}).
As we can see from 
that Table,
although the task is harder, using the fine-grained types for training results in slightly better performance on identifying non-referring markables in general than models trained on a single {\NR} class. 
In term of the performance on coreference chains, the {\FINENR} approach achieved the same score 
as the {\PREFILTERING} approach and slightly lower than the {\HYBRID} approach (see Table \ref{table:finalcracaccuracy}).

\textbf{Training without Singletons and Non-referring} 

Finally, we trained our model without singletons and non-referring expressions ({\NONR}) to assess their effects on non-singleton clusters (i.e. the standard {\CONLL} setting). Since here we evaluate in a singleton excluded setting, we report for our models trained with singletons and non-referring expressions the standard {\CONLL} scores with singletons and non-referring markables excluded.  
As shown  
in Table \ref{table:finalcracaccuracy}, all three models trained with additional singleton and non-referring markables achieved better {\CONLL} scores when compared with the newly trained model. 
The system achieves substantial  
gains of up to 1.4 percentage points ({\HYBRID}) by training with the additional singletons and non-referring expressions. This suggests that the availability of singletons and non-referring markables can help the decisions made for non-singleton clusters.

\textbf{State-of-the-art Comparison} Since the {\CRAC} corpus was released recently, the only published results are those by the baseline system \cite{lee-et-al:CL13} on the shared task \cite{poesio2018crac}. Our best system ({\HYBRID}) outperforms this baseline by large margins (5.8\% and 13.4\% when evaluated with or without singletons respectively) (see Table \ref{table:finalcracaccuracy}) even though that system was evaluated on gold mentions.

\subsection{Evaluation on the {\PD} data set}
We then test our best system on the {\PD} corpus\footnote{\newcite{poesio-etal-2019-crowdsourced} uses an early version of our system.}. We compare our system with the results by \newcite{poesio-etal-2019-crowdsourced} (Table \ref{table:finalpdaccuracy}). Our system is 3\% better when evaluated with singletons included and outperforms their system by 5.3\% when evaluated without the singletons. In addition, our system achieved an F1 of 56.7\% on non-referring expressions and this is 2.1\% better than their result (54.6\%). Overall, our system achieved the new SoTA on the {\PD} data.

\subsection{Evaluation on the {\CONLL} data set}

Finally, we tested our models on 
the {\CONLL} data to assess the performance of our system on the standard data set. 
Table \ref{table:finalaccuracy} compares our results with those of the top-performing systems on {\CONLL} at the present time. 
We report precision, recall and F1 scores for all three major metrics (MUC, B$^3$ and CEAF$_{\phi_4}$) and mainly focus on the average {\CONLL} F1 scores presented in the last column. As showed in Table \ref{table:finalaccuracy}, our model achieved a {\CONLL} score of 76.4\%, which is only 0.2\% lower than the best-reported result at present, 
achieved by 
\cite{kantor2019bertee} that use a similar mention representations as 
our system. 
Although the systems by \newcite{joshi2019bert} and \newcite{joshi2019spanbert} have better results than the \newcite{kantor2019bertee} system, it is not directly comparable with our system, as their systems are fine-tuned on BERT. 
Such systems need to be trained on GPUs with 32GB memory, which are not available to our group. 
By contrast, our system was trained with 
a GTX 1080Ti GPU with %that has
an 11GB memory.

\begin{table}[t]
\centering
\small
\begin{tabular}{llllll}
\toprule
&\multicolumn{5}{c}{Positions}\\\cmidrule{2-6}
Size&1&2&3&4&5-7\\\midrule
2&0.55&0.45\\ 
3&0.38&0.32&0.29\\
4&0.29&0.24&0.23&0.22\\
5&0.24&0.20&0.19&0.19&0.19\\
6&0.19&0.17&0.16&0.17&0.15\\
7&0.18&0.14&0.14&0.14&0.13\\
\bottomrule
\end{tabular}
\caption{\label{table:cluster_att_eval} The average mention importance %salience 
attention scores in the {\CONLL} development set, grouped by  mentions position and cluster size in the final clusters. }
\end{table}

\begin{table}[t]
\centering
\small
\begin{tabular}{l l l}
\toprule
&Avg. F1& $\Delta$\\
\midrule
Our model&76.9&\\
%\ \ + C2f&76.1&0.8\\
\ \ - Position emb&76.2&0.7\\
\ \ - Width emb&76.5&0.4\\
\ \ - Cluster history&75.9&1.0\\
\ \ - Oracle cluster&76.3&0.6\\
\bottomrule
\end{tabular}
\caption{\label{table:analysis} The comparison between our best model and different ablated models on {\CONLL} development set.}
\end{table}

\subsection{Discussion}

We further 
analyze %analysis 
our model on the {\CONLL} data to give a more detailed study on different aspects of our model. 
(We use the standard {\CONLL} data instead of the {\CRAC} data because the {\CONLL} corpus is larger than the {\CRAC} corpus and is widely used. 
As a result, the analysis on {\CONLL} data might also be beneficial for 
other researchers focusing on {\CONLL} only.) 

\textbf{Mention Importance} 
We first
assess our hypothesis that our attention scores can capture mention importance--i.e., the finding from the linguistic and psychological literature on anaphora that the initial mentions of an entity %are those 
tend to include more information, whereas the following mentions are generally reduced. 
Table \ref{table:cluster_att_eval} shows an analysis of the attention scores that supports this hypothesis. 
We computed the average attention scores for mentions in a cluster in order of mention. 
Clusters that have different size are analysed separately, as scores from different-sized clusters are not directly comparable. 
As we can see from the Table, after analysis the attention scores assigned to the mentions at  different positions in the cluster, we find that the attention scores assigned to the first mention in a cluster are always higher than others, which is in line with linguistic findings that mentions introducing an entity are more informative. 
This suggests that our attention model does capture something like mention importance.

\textbf{Why Cluster Ranking?} The reason why we use a cluster ranking approach instead of  mention ranking is not only because it is linguistically more appealing, but also due to several practical restrictions of the mention ranking models. First of all, the current SoTA mention-ranking systems tend to be hybrids, using  entity-level features alongside 
mention-pair features. 
Thus, such models 
are usually more complex 
than pure mention ranking models, 
and substantially
increase the number of trainable parameters. 
Take \newcite{lee2018higher} system as an example.
The mention ranking part of the system contains 4.8M parameters, but the full system has double the number of parameters  (9.6M) to access entity-level features. 
Our system, on the other hand, links the mentions directly to the entity and uses only 4.8M parameters, which is much simpler than such hybrid  models. 
Second, we hope that using a cluster ranking model 
will allow us to explore rich cluster level features and advanced search algorithms (e.g. beam search) in future work. 

\textbf{The Effect of Oracle Clusters on Training Time} 
Training  cluster ranking systems using system clusters is time-consuming:
Our model trained on  system clusters takes 80 hours to train for 200K steps, which is much more than 
the 48 hours training time of the 
\newcite{lee2018higher} system (400K steps). 
The main reason the cluster ranking system 
is slower than
its mention ranking counterpart is that the cluster ranking model processes one mention at a time, hence does not benefit from parallelization. 
To solve this problem, we trained the system on  oracle clusters instead.
The oracle clusters are created by using the system mentions with the gold cluster ids.
By doing so all the clusters can be created before resolving the mentions into the entities. 
As a result, the training (200K steps) can be finished in as little as 16 hours, which is 5x faster than training the model on system clusters, and 3x faster than training the mention ranking model.

\subsection{Ablation study}
We removed different parts %part 
of our model to show the importance of the individual part of our system (see Table \ref{table:analysis}).

\textbf{Position Embeddings} We first removed the position embeddings,  used in the self-attention to determine the relative importance of the mentions in the cluster. 
By removing the position embeddings, the relative importance of a mention becomes independent of its position in the cluster. As a result, the performance of the model drops by 0.7\%.

\textbf{Width Embeddings} We then removed the cluster width embeddings from our features. The cluster width embedding is a feature used in computing the pairwise scores, which allows mentions to known the size of individual candidate clusters. (Cluster size can be used as an indicator of  cluster salience, as the larger the size, the more frequently an entity is mentioned, having therefore a higher salience.)
The cluster width feature contributes 0.4\% towards our model.

\textbf{Cluster History} We trained a model that keeps exactly one cluster per entity, and
the history clusters are excluded from the candidate lists. 
This removing of history clusters reduces the chance of linking the mentions to the correct entity; as a consequence, the performance drops by 1 percentage point. 

\textbf{Oracle Clusters} Finally, we trained a model using the system clusters directly instead of the oracle clusters. 
As we mentioned in the previous section,  training on the system clusters is more time consuming than training on the oracle clusters. 
And replacing these clusters  suggests that training on the oracle clusters is not only faster, but also results in  better performance (0.6\%).

\section{Related Work}
\textbf{Pure Mention Ranking Models} Most recent coreference systems are highly reliant on  mention ranking, which is
effective and generally faster to train  
compared with the cluster ranking system. 
Systems  based only on the mention ranking model include \cite{wiseman2015learning,clark2016improving,lee2017end}. \newcite{wiseman2015learning} introduced a neural network based approach to solve the task in a non-linear way. 
In their system, the heuristic features %that
commonly used in  linear models are transformed by a $\tanh$ function to be used as the mention representations. 
\newcite{clark2016improving} integrated  reinforcement learning to let the model optimize directly on the B$^3$ scores. 
\newcite{lee2017end} first presented a neural joint approach for mention detection and coreference resolution. Their model does not rely on  parse trees; instead, the system learns to detect  mentions by exploring the outputs of a BiLSTM.

\textbf{Models using Entity Level Features}
Researchers have been aware of 
the importance of  entity level information 
at least since 
\newcite{luo-et-al:ACL04}, and many systems trying to  exploit  cluster based features have been proposed since.
Among neural network models,
\newcite{bjorkelund2014learning} built a latent tree system that explores non-local features through beam search. 
The global feature-aided model showed clear gains when compared with the model based only on pairwise features. 
\newcite{clark2015entity} introduced a entity-centric coreference system by manipulating the scores of a mention pair model. 
The system first runs a mention pair model on the document and then uses an agglomerative clustering algorithm to build the clusters in an easy-first fashion.
This system was later extended by \newcite{clark2016improving} to make it run on neural networks. 
\newcite{wiseman2016learning} add to the \newcite{wiseman2015learning} system an LSTM to encode the partial clusters. 
The outputs of the LSTM are used as additional features for the mention ranking model. 
\newcite{lee2018higher} 
is an extended version of \newcite{lee2017end} mainly  enhanced by using ELMo embeddings \cite{peters2018elmo}, 
but the use of  second-order inference enabled the system explore partial entity level features and further improved the system by 0.4 percentage points. 
Later the model was further improved by \newcite{kantor2019bertee} who use  BERT embeddings \cite{devlin2019bert} instead of ELMo embeddings. 
At this stage, both BERT and ELMo embeddings are used in a pre-trained fashion. 
Recently, \newcite{joshi2019bert} fine-tunes the BERT model for coreference task, result in again a small improvement. Later, \newcite{joshi2019spanbert} introduces a BERT model (SpanBERT) specifically trained for the tasks that involves spans, by using the SpanBERT, the system achieved a substantial gain of 2.7\% when compared with the \newcite{joshi2019bert} model.

\textbf{Cluster Ranking Models} To the best of our knowledge, our system is the only recent system that does \textit{not} rely on a mention ranking model. 
However, there are a number of early studies that laid a solid foundation for the cluster ranking models (see \cite{poesio-et-al:ana_book_heuristic} for a survey).
The best known `modern' examples are the systems proposed by \newcite{luo-et-al:ACL04} and by \newcite{rahman&ng:JAIR11}, 
but this approach 
was the dominant model for anaphora resolution at least until the paper by \newcite{soon-et-al:CL01}, as it directly implements the linguistically and psychologically motivated view that anaphora resolution involves the creation of a discourse model articulated around discourse entities \cite{karttunen:76}. 
The entity mention model of \newcite{luo-et-al:ACL04} introduced the notion that a training instance consists of a mention and an active cluster, and therefore allowed for cluster-level features encoding information about multiple entities in the cluster. 
\newcite{luo-et-al:ACL04} also proposed a clustering algorithm in which the clustering options are encoded in a Bell tree that also specifies the coreference decisions resulting in a cluster--an idea related to our idea of cluster history.
\newcite{rahman&ng:JAIR11} introduced the term `cluster ranking' and greatly developed the approach, e.g., by introducing a rich set of cluster-level features.
Their model was the first cluster-ranking model to significantly outperform mention pair models. 

\textbf{Singletons and Non-referring Expressions} 
Again, to the best of our knowledge, ours is the only modern neural network-based, full coreference system that attempts to output singletons and non-referring markables.
The Stanford Deterministic Coreference Resolver \cite{lee-et-al:CL13} uses a number of filters to \textit{exclude} expletives as well as quasi-referring mentions such as percentages (e.g., \textit{9\%}) and measure {\NP}s (e.g., \textit{a liter of milk}) and its extension proposed by \newcite{DeMarneffe:2015:MLD:2831407.2831417} includes more fiters to exclude singletons, but these aspects of the system are not evaluated. 
The best-known systems also attempting  to  annotate 
non-referring markables date back to the pre-{\ONTONOTES} era. 
The pronoun resolution algorithm proposed by \newcite{lappin&leass:94} includes a series of hand-crafted heuristics to detect expletives.
The statistical  classifier proposed by \newcite{evans2001applying} classifies pronouns in several categories which, apart from nominal anaphoric, include cataphoric, pleonastic, and clause-anaphoric.
\newcite{versley-EtAl:2008:PAPERS} used the BBN pronoun corpus to confirm the hypothesis that tree kernels would be well-suited to identify expletive pronouns. 
\newcite{boyd-et-al:05} develop a set of hand-crafted heuristics to identify non-referring \textit{nominals} in the sense of \newcite{karttunen:76}. 
The systems developed by Bergsma and colleagues to identify pronominal \textit{it} with a classifier using a combination of lexical features and web counts \cite{bergsma-lin-goebel:2008:ACLMain,Bergsma:2011:nada}. 
A lot of work on identifying expletives was carried out in the context of the DiscoMT evaluation campaigns, but this work was typically only focused on disambiguating pronoun \textit{it} \cite{loaiciga-et-al:EMNLP17}.
For more discussion of these and other systems, see \cite{Uryupina-et-al:ana_book_non_referring}.

\section{Conclusions}
In this work, we presented the first neural network based  system for full coreference resolution also covering singletons and non-referring markables. 
Our system uses an attention mechanism to form the cluster representations using mention importance scores from the mentions belonging to the cluster. 
By training the system on  oracle clusters we show that a cluster ranking system can be trained 5x faster, and faster than a mention-ranking system with a similar architecture. 
Evaluation on the {\CRAC} corpus shows that our system is 5.8\% better than the only existing comparable system, the Shared Task baseline system that used the gold mentions. The evaluation on {\PD} shows the same trend. 
Further evaluation on the {\CONLL} corpus shows our system achieves on that corpus, for the subtask in which singleton and non-referring expression detection are excluded, a performance equivalent  to that of the SoTA \newcite{kantor2019bertee} system.
We also demonstrated that a large improvement on non-singleton coreference chains can be made by training the system with additional singletons and non-referring expressions.

\section{Acknowledgments}

This research was supported in part by the DALI project, ERC Grant 695662.

\newpage

% \nocite{*}
\section{Bibliographical References}
\label{main:ref}

\bibliographystyle{lrec}
\bibliography{lrec2020}

%\section{Language Resource References}
%\label{lr:ref}
%\bibliographystylelanguageresource{lrec}
%\bibliographylanguageresource{lrec2020W-xample}

\end{document}